# K-Algorithm: A Modified Technique for Noise Removal in Handwritten Documents

Kanika Bansal[1], Rajiv Kumar[2]

[1]Student, SMCA, Thapar University, Patiala, Punjab, India.

knbs_ind@yahoo.com

[2]Assistant Professor, SMCA, Thapar University, Patiala, Punjab, India.

rajiv.patiala@gmail.com

## ABSTRACT

OCR has been an active research area since last few decades. OCR performs the recognition of the text in the scanned document image and converts it into editable form. The OCR process can have several stages like preprocessing, segmentation, recognition and post processing. The preprocessing stage is a crucial stage for the success of OCR, which mainly deals with noise removal. In the present paper, a modified technique for noise removal named as "K-Algorithm" has been proposed, which has two stages as filtering and binarization. The proposed technique shows improvised results in comparison to median filtering technique.

## KEYWORDS

*OCR, Handwritten, Filtering, Binarization, Preprocessing, Offline.*

## 1. INTRODUCTION

The offline handwritten character recognition refers to the processing of the text images captured with the help of scanner, which gives meaningful symbols as output, pertaining to the script. The counterpart, online handwritten character recognition deals with automatic conversion of the characters, which are written on a special digitizer, tablet PC or personal digital assistant (PDA) where a sensor picks up the pen-tip movements as well as pen-up/pen-down switching. The online character recognition captures the dynamic motion during handwriting. The offline handwritten character recognition process possesses more complexity than the online handwritten character recognition process, attributed to the fact that more information may be captured in the online case such as the direction, speed and the order of strokes of the handwriting. Since this paper considers the offline handwritten character recognition process, it is discussed in detail.

The offline images are often obtained through scanning of the text documents. The scanned image, thus obtained, can only be read and not be edited but there may be applications where the text from the scanned image needs to be edited. Here, the character recognition process offers a solution, which translates the image into editable form that can be processed by the computer. This process can be used to translate articles, books and documents into electronic format, to publish text on website, to process the cheques in banks, to sort letters and many more applications.

### 1.1. Stages of Character Recognition Process

The image acquired through scanner needs to pass through various stages which can be named as preprocessing, segmentation, recognition and post processing as given by [1,2]. The various stages of the offline handwritten character recognition process are shown in Figure 1.

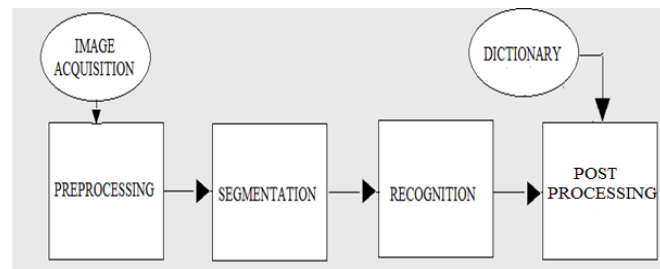

Figure 1. Stages of Character Recognition Process

*Preprocessing Stage*: The acquired image is required to go under a number of steps to convert it to a usable form for further stages of recognition process. This constitutes the preprocessing stage, which includes all the functions required to produce the cleaned up version of image acquired through scanner. It includes the processes like filtering, binarization, thinning, smoothing.

*Segmentation Stage*: The cleaned up image obtained as output of preprocessing stage serves as input to the segmentation stage, which refers to decomposing the document into subcomponents. It deals with the separation of lines, words and characters according to the application through usage of various segmentation strategies.

*Recognition Stage* :The segmented image now act as input for the recognition stage which includes the various pattern recognition strategies to assign an unknown sample to a predefined class to identify the character according to the script used.

*Postprocessing Stage*: The recognized characters are fed into the post processing stage, which includes the usage of dictionary according to the script for purpose of spell check, grammar check in order to enhance the rate of recognition and minimize the recognition errors.

The above stages constitute the character recognition process and since this paper deals with preprocessing, it is further discussed in length in the following section.

## 2. PREPROCESSING AND ITS SIGNIFICANCE

The image acquired with the help of scanner may contain many impurities, which may be introduced due to different reasons like varying paper quality, varying ink quality and the quality of scanner used. Therefore, the acquired image cannot be sent directly to the subsequent stages of recognition process and needs the impurities to be removed, which is done in the preprocessing stage.

Preprocessing consists of a series of steps, which are used to produce a cleaned-up version of the acquired image. Preprocessing converts the acquired image into a more usable form for the next stages. The preprocessing stage plays an important role in the handwritten character recognition process. The extent to which acquired image is appropriately cleaned, influences the rate of accuracy of segmentation and hence recognition. Inappropriate preprocessing increases the complexity of techniques used in further stages and may lead to incorrect segmentation and recognition. The major objectives of preprocessing stage can be to reduce the amount of noise present in the document and to reduce the amount of data to be retained. The preprocessing stage includes a number of techniques in order to achieve these objectives, which are now explained.

## 2.1. Reduction of Noise

Noise refers to the error in the pixel value or an unwanted bit pattern, which do not have any significance in the output. Noise may be introduced due to reproduction and transmission of image during its acquisition process. Noise may introduce gaps in lines, disconnected segments and filled loops in the document. The salt and pepper noise is generally encountered in the poor quality documents and it appears as isolated pixels or as on pixels on off region or as off pixels in on region. This noise can be removed through filtering as given in [3].

Filtering refers to the usage of a predefined function with the image in order to assign a value to a pixel in consideration as a function of the values of its neighbouring pixels. It diminishes unwanted bit patterns, introduced by uneven writing surface or poor quality of the data acquisition device. It also removes slightly textured or coloured background and sharpens the image. The objective in the design of a filter to reduce noise is that it remove as much of the noise as possible while retaining all of the relevant data. The filters can be categorised into the linear filters and non-linear filters. The widely used linear filters like mean filter and wiener filter are often used to remove noise but they have several disadvantages like blurring of edges and fine details, destruction of lines. Thus, non-linear filters like median filters are used which overcomes the limitations of linear filters to an extent. As stated in [4], the median filter too causes the removal of corners and threads and may cause blurring of text, thus to overcome these limitations, a modified technique is developed, which is discussed in the following section.

## 2.2. Data Reduction

It basically refers to the reduction in the amount of data, which is to be retained in the preprocessed image. It encodes the information using the fewer bits than the original representation would use. It reduces the consumption of expensive resources like memory and time. It is often achieved through binarization. Binarization refers to the conversion of a gray scale image into binary image where a pixel can have any of the two values (0 or 1). This process works on the basis of threshold value. This threshold value can be decided locally and globally as stated in [5]. The global threshold picks one value for entire image on the basis of estimation of the background intensity from the intensity histogram of the image whereas the local threshold picks different values for each pixel on the basis of the local information as stated in [6]. The filtering technique described above is modified and combined with the binarization technique to form a modified technique, which is described in the following section.

## 3. K - ALGORITHM : A MODIFIED TECHNIQUE FOR NOISE REMOVAL

As discussed in section 2, the median filter overcomes the limitations of the linear filters, but it has its own disadvantages like it may cause the removal of corners and threads, blurring of text in the document. In order to overcome these limitations, K-Algorithm, which is the modified technique for the removal of noise in the handwritten documents, is proposed in this paper.

The proposed solution consists of two steps (i) filtering and (ii) binarization.

### STEP I

In this step, the filtering technique is used for removal of noise. This step uses the median filter as its basis. The median filter ranks the neighbouring pixels according to their intensity and the median value becomes the new value for the central pixel. The median filter operation can be seen in the Figure 2.

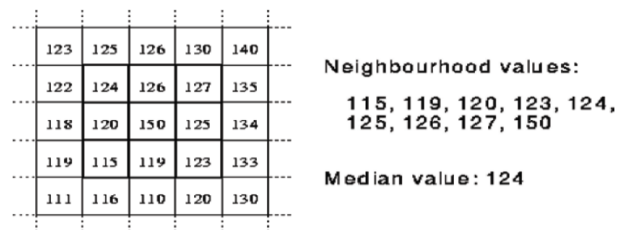

Figure 2. Median Filter Example

The median filter serves as the basis for the modified technique. The modified technique uses the median filter conditionally, that is, the application of the median filter to a pixel depends on the parameter, 'K' and hence, the algorithm derives its name, "K-Algorithm". The parameter, 'K' denotes the count of the lowest intensity pixels in the current neighbourhood matrix of the pixel under consideration. If the value of parameter, K is less than defined value, only then the median filtering is applied to the current pixel. The defined value depends on the matrix size used. The output of the filtering may contain some textured or slightly coloured background, which can be removed during binarization process in step II, to be discussed in the following section.

## STEP II

The output of the filtering technique may still have some textured or slightly coloured background, which may interfere in the functioning of the subsequent stages. In order to deal with this, a binarization technique is used to convert the filtered image to a binary image. In this technique, a threshold value is calculated and the pixels having intensity value above it are set to white(0) and the pixels having the intensity below it are set to black(1). The threshold value is calculated by taking the average of all pixel intensities in the document as given in [7].

The formal algorithm for the above solution is given below:-

Parameters used:

Image:-The bitmap image given as input.

Matrix_Size:-Defines the neighbourhood matrix size for the pixel. In case of 3x3 matrix, it has value as two.

Min_X:-Defines the minimum x coordinate value for input image.

Max_X:-Defines the maximum x coordinate value for input image.

Min_Y:-Defines the minimum y coordinate value for input image.

Max_Y:-Defines the maximum y coordinate value for input image.

Pixel_Intensity(X, Y):-Returns or Sets the pixel intensity at coordinates X and Y.

Pixel_Values:-A list storing intensity values of all pixels.

K:-Parameter defined according to matrix size. For 3x3 matrix, it has value as one.

Pixel_Values_Count:-Defines count of values in list Pixel_Values.

**-- Step I: Filtering**

Modified_Median_Filter (Image, Matrix_Size)

Set A_Min=-(Matrix_Size)/2

```
Set A_Max= (Matrix_Size)/2
For X=Min_X to Max_X                                                //1
    For Y=Min_Y to Max_Y                                            //2
        For X1=A_Min to A_Max                                       //3
            Set Temp_X=X+X1
            If (Temp_X>=Min_X and Temp_X<=Max_X)
                For Y1=A_Min to A_Max                               //4
                    Set Temp_Y=Y+Y1
                    If (Temp_Y>=Min_Y and Temp_Y<=Max_Y)
                        Add Pixel_Intensity (Temp_X, Temp_Y) to
                        list Pixel_Values
                    End If
                End For                                             //4
            End If
        End For                                                     //3
        Sort the list Pixel_Values
        Set No_Occurences=Number of the occurrences of lowest pixel intensity value
        in list Pixel_Values
        If (No_Occurences==K)
            Median_value=Value at Pixel_Values_Count/2
            Set Pixel_Intensity(X, Y) =Median_Value
        End If
    End For                                                         //2
End For                                                             //1
Return Modified Image
```

## -- Step II: Binarization

```
Binarization (Image)
For X=Min_X to Max_X                                                //1
    For Y=Min_Y to Max_Y                                            //2
        Pixel_Intensity_Sum=Pixel_Intensity_Sum+Pixel_Intensity(X, Y)
        Pixel_Count=Pixel_Count+1
    End For                                                         //2
End For                                                             //1
Average_Intensity= Pixel_Intensity_Sum/ Pixel_Count
For X=Min_X to Max_X                                                //3
```

```
        For Y=Min_Y to Max_Y                                            //4
            If (Pixel_Intensity(X, Y) >=Average_Intensity)
                Set Pixel_Intensity(X, Y) =WHITE
            Else
                Set Pixel_Intensity(X, Y) =BLACK
            End If
        End For                                                         //4
End For                                                                 //3
Return Modified Image
```

The proposed algorithm is implemented and results are discussed in the following section.

## 4. EXPERIMENTAL RESULTS AND DISCUSSION

The proposed algorithm has been implemented by the authors in the .NET framework using C# language.The Drawing namespace and Bitmap class present in the inbuilt libraries of C# is utilised to process the bitmap image. The image acquired through scanner is shown in Figure 3. It can be seen that the noise is present in the scanned document and thus, it requires preprocessing.

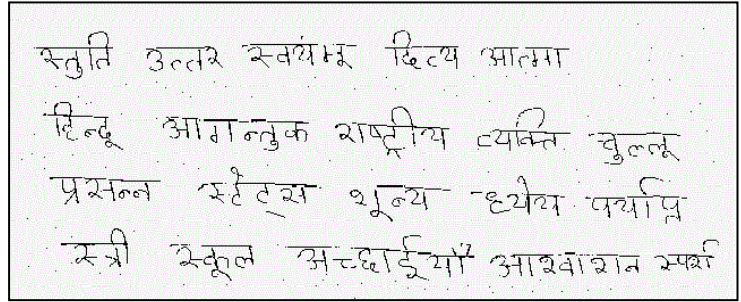

Figure 3. Original Noisy Image

Initially, the median filter is used to remove the noise from the scanned image. The Figure 4 shows the results after application of median filter. It can be seen that although, noise has been removed from scanned image, but the text got blurred in the resultant image and it is not suitable for processing by subsequent stages of handwritten character recognition process. Hence, it requires another approach for preprocessing.

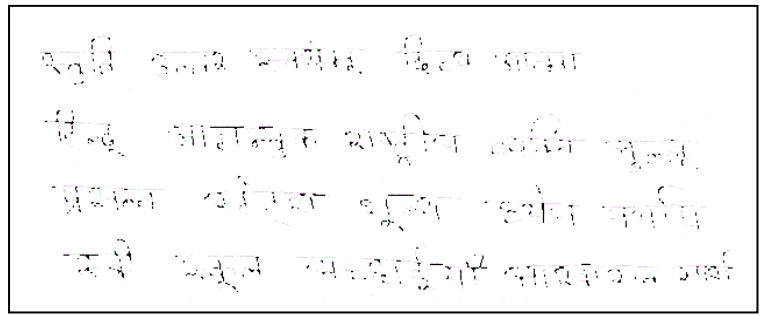

Figure 4. Resultant Image after application of Median Filter

The modified technique-"K-Algorithm" is then applied to the scanned image. The step I of algorithm, which is filtering is executed and the resultant image is shown in Figure 5. It can be observed that the text has not been blurred as in Figure 4 and the resultant image shows improvement in clarity of relevant pixels. There is presence of slightly textured background in the image as seen in Figure 5. Since, this can interfere in the accurate working of subsequent stages; it needs to be removed and can be removed through step II of the modified technique-"K-Algorithm".

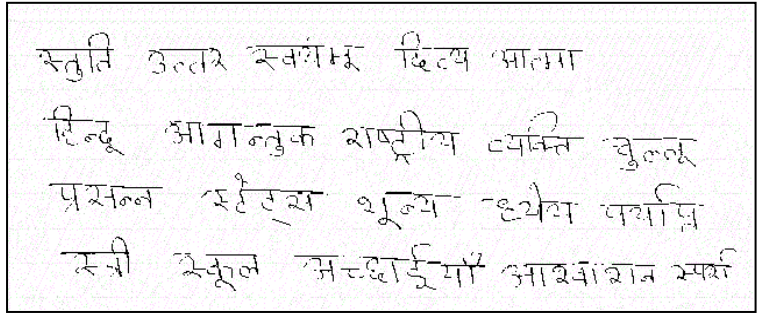

Figure 5. Resultant Image after application of Modified Technique

The step II of the modified technique-"K-Algorithm", which is binarization, is applied to resultant image of step I and the resultant image is shown in Figure 6. It can be observed that the textured and coloured background present in the Figure 5 is now removed and it is suitable to be served as input to further stages of handwritten character recognition process.

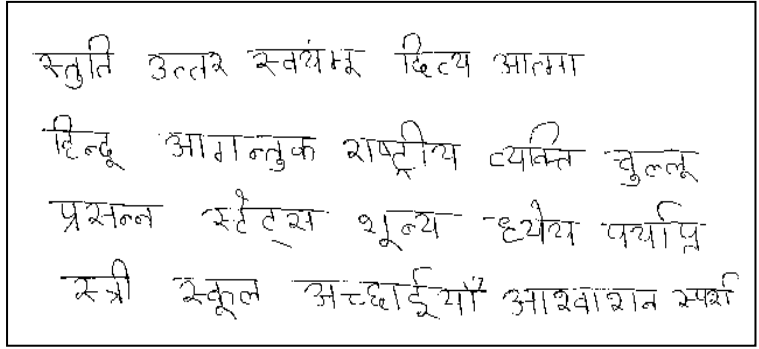

Figure 6. Resultant Image after application of Binarization

## 5. CONCLUSION

The "K-Algorithm" has been implemented and tested on several documents. It can be seen from the results that the modified technique provides much better preprocessing than the existing techniques like median filtering. The modified technique is able to overcome the disadvantages of the existing techniques. The results obtained are good and encouraging. The above work highlights the significance of the preprocessing stage in the handwritten character recognition process.

[3] R. Kannan, R. Prabhakar and R. Suresh, (2008) "Off-Line Cursive Handwritten Tamil Character Recognition," in Proceedings of International Conference on Security Technology, pp. 160-164.

[4] T. Saba, G. Sulong and A. Rehman, (2010) "Document Image Analysis: Issues, Comparison Of Methods And Remaining Problems". Springer Science+Business Media B.V., pp. 101-110.

[5] S. Mo and J. Mathews, (1998) "Adaptive, Quadratic Preprocessing of Document Images for Binarization," IEEE Transactions on Image Processing. vol. 7, no. 7, pp. 992-999.

[6] R. Kasturi, L. Gorman and V. Govindaraju, (2002) "Document Image Analysis: A Primer". S¯adhan¯a, vol. 27, part 1, pp. 3–22.

[7] R. Kumar and A. Singh, (2011) "Algorithm to Detect and Segment Gurmukhi Handwritten Text into Lines, Words and Characters" IACSIT International Journal of Engineering and Technology, vol. 3, no.4.

**AUTHORS**

Rajiv Kumar obtained his Ph.D. from the University College of Engineering, Punjabi University, Patiala, Punjab, India. At present, he is the Assistant Professor in the School of Mathematics and Computer Applications, Thapar University, Patiala. He has published several articles in international journals and conferences.

Kanika Bansal obtained her B.Tech. (CSE) from Guru Gobind Singh Indraprastha University, Delhi, India and currently, pursuing her M.Tech. from Thapar University, Patiala, Punjab, India. She is currently working in the document image processing area.